\documentclass[11pt]{article}
\usepackage[margin=1in]{geometry}
\usepackage[T1]{fontenc}
\usepackage{lmodern}
\usepackage{microtype}
\usepackage{amsmath,amssymb}
\usepackage{booktabs}
\usepackage{tabularx}
\usepackage{array}
\usepackage{graphicx}
\usepackage[numbers,sort&compress]{natbib}
\usepackage[hidelinks]{hyperref}
\usepackage{url}
\usepackage{enumitem}
\usepackage{caption}
\usepackage{subcaption}
\usepackage{float}
\newcolumntype{Y}{>{\raggedright\arraybackslash}X}

\title{\textbf{SNOMED CT Concept Recommendation from Masked Clinical Context}\\
\large Sparse, Dense, and Retrieval-Augmented Approaches}
\author{Ali Noori\\
Informatics and Analytics, University of North Carolina Greensboro\\
Greensboro, North Carolina, USA}
\date{}

\begin{document}
\maketitle

\begin{abstract}
Mapping clinical language to SNOMED CT is central to semantic interoperability, clinical analytics, and reusable phenotyping, yet concept recommendation becomes difficult when the target mention is absent from the observed text or poorly represented in training data. We present a masked-concept recommendation benchmark using the SNOMED CT Entity Linking Challenge v1.2.1 release derived from MIMIC-IV-Note. The analyzed release contains 75,491 annotations across 272 discharge summaries. We reconstruct the historical 204-note training and 68-note test partitions, collapse repeated mentions to one note-concept pair, remove the target concept mention from a local context window, and rank candidate concepts observed in training. The historical test partition contains 14,179 unique note-concept pairs, of which 12,809 (90.34\%) have targets represented in the training vocabulary and form the primary closed-vocabulary evaluation set. We compare popularity, sparse TF-IDF concept prototypes, 96-dimensional latent semantic analysis (LSA), sparse-dense fusion, similarity-weighted retrieval from training notes, and a retrieval-augmented hybrid. Sparse TF-IDF provides the strongest performance, achieving Recall@1 of 14.81\%, Recall@10 of 33.43\%, mean reciprocal rank (MRR) of 0.2114, and nDCG@10 of 0.2297. Dense LSA reaches 17.10\% Recall@10 and 0.0735 MRR, while the retrieval-augmented hybrid reaches 31.99\% Recall@10 and 0.1937 MRR. Paired note-cluster bootstrap analysis shows that the retrieval-augmented hybrid is lower than sparse TF-IDF for Recall@10 (95\% CI of the difference, -0.0184 to -0.0104) and MRR (-0.0212 to -0.0142). Performance is strongly frequency-dependent: sparse TF-IDF Recall@10 is 7.74\% for concepts appearing in one or two training notes and 43.90\% for concepts appearing in more than ten. In addition, 1,370 test note-concept pairs (9.66\%) contain targets absent from the training vocabulary. These findings establish a reproducible retrieval baseline and show that, in this low-resource setting, terminology coverage and local lexical context are more consequential than simple whole-note retrieval augmentation. The results motivate terminology-level retrieval with biomedical encoders and ontology-grounded candidate generation as the next stage.
\end{abstract}


\section{Introduction}
Clinical notes contain diagnoses, symptoms, procedures, measurements, observations, and clinician reasoning that are difficult to reuse computationally without normalization. Automated clinical coding and entity linking map this free text to standardized identifiers, supporting interoperability and reducing dependence on manual abstraction \citep{dong2022coding}. SNOMED CT is particularly important because it provides a large, structured clinical terminology for detailed representation of healthcare information \citep{donnelly2006snomed}.

The SNOMED CT Entity Linking Challenge created an important resource for clinical terminology mapping by pairing MIMIC-IV-Note discharge summaries with curated SNOMED CT annotations \citep{davidson2025challenge,hardman2026challenge}. The original competition focused on locating entity spans and assigning concept identifiers. The PhysioNet v1.2.1 release subsequently incorporated the historical hidden test set and quality-control updates \citep{hardman2026challenge}. This resource also exposes a central difficulty of clinical concept modeling: the target terminology is large and highly long-tailed, so many concepts have very few supervised examples.

Our prior work has addressed complementary parts of this problem. We examined robust evaluation strategies for automated ontology annotation systems \citep{noori2025metrics}, evaluated a Bi-GRU sequence-labeling approach for SNOMED CT concept annotation in clinical text \citep{noori2025bigru}, and later showed that SNOMED CT co-occurrence and embedding-based semantic similarity capture complementary relationships in clinical documentation \citep{noori2026semantic}. Together, these studies motivate a shift from detecting and describing concepts toward recommending clinically plausible concepts from context.

Retrieval-augmented generation (RAG) provides a general framework for grounding model outputs in external evidence \citep{lewis2020rag}. Dense retrieval can recover semantically related evidence even when exact lexical overlap is limited \citep{karpukhin2020dpr}, and biomedical RAG studies increasingly emphasize the importance of retriever design, terminology knowledge, and clinically appropriate evaluation \citep{liu2025ragreview}. Recent systems have also applied retrieval and large language models to SNOMED CT mapping \citep{huh2025mapping,ehrler2026rag}, while ontology-grounded retrieval explicitly uses structured knowledge to guide evidence selection \citep{sharma2025ograg}.

The present study asks whether retrieval augmentation already improves concept recommendation when only the credentialed challenge notes and annotations are used. We formulate recommendation as masked concept recovery: the observed target mention is removed from a local clinical context, and the system must rank the missing SNOMED CT concept. This design prevents direct string matching and isolates the information available from surrounding clinical language.

We address three research questions:
\begin{enumerate}[leftmargin=1.4em,label=\textbf{RQ\arabic*:}]
    \item How accurately can sparse and dense context representations recover a masked SNOMED CT concept?
    \item Does evidence retrieved from training notes improve ranking beyond the strongest context-only baseline?
    \item How strongly are recommendation performance and attainable coverage affected by concept frequency and training-vocabulary overlap?
\end{enumerate}

The study makes four contributions. First, it defines a reproducible masked-concept recommendation benchmark from the SNOMED CT Entity Linking Challenge v1.2.1 release. Second, it compares sparse TF-IDF prototypes, data-derived dense LSA embeddings, sparse-dense fusion, retrieved-note evidence, and retrieval-augmented ranking under a leakage-controlled historical split. Third, it reports vocabulary coverage and frequency-stratified performance rather than presenting a single aggregate score. Fourth, it uses paired bootstrap resampling clustered by note to quantify uncertainty in the comparison between retrieval augmentation and the strongest sparse baseline.

The term \emph{retrieval-augmented} in this manuscript refers to adding non-parametric evidence retrieved from training notes to the ranking function. We do not evaluate a generative LLM in this experiment. The challenge archive does not contain the complete SNOMED CT terminology descriptions, synonyms, hierarchy, and relation files needed for terminology-level ontology-grounded RAG. The present study therefore evaluates the retrieval and ranking layer and uses the results to specify what a stronger biomedical-encoder and ontology-backed system must improve.

\section{Related Work}
\subsection{Clinical coding, ontology annotation, and SNOMED CT entity linking}
Automated clinical coding has been studied with rules, dictionaries, statistical models, neural architectures, and language models \citep{dong2022coding}. MedCAT combines concept recognition and contextual disambiguation across clinical domains \citep{kraljevic2021medcat}. In our earlier clinical annotation work, a Bi-GRU sequence-labeling model was evaluated for identifying SNOMED CT concept spans in clinical text, illustrating the use of supervised neural models for terminology annotation \citep{noori2025bigru}. Separately, our work on automated ontology annotation examined evaluation metrics for comparing annotation systems, motivating the use of multiple complementary measures rather than a single headline score \citep{noori2025metrics}.

The SNOMED CT Entity Linking Challenge substantially expanded the available benchmark resource, with the competition paper reporting 74,808 curated annotations across 272 discharge summaries \citep{davidson2025challenge}. The v1.2.1 PhysioNet release reflects later quality-control changes \citep{hardman2026challenge}. Conventional entity linking is given the original note and seeks the entity span and concept ID. In contrast, the present task uses an annotated concept instance only to construct evaluation data, masks the target surface form, and asks a recommender to rank the missing concept from surrounding context. It therefore evaluates context-driven suggestion rather than span detection.

\subsection{Sparse and dense retrieval}
Sparse vector-space methods remain strong information-retrieval baselines because local lexical context can contain highly discriminative evidence. TF-IDF weighting emphasizes terms that distinguish documents or contexts within a collection \citep{salton1988tfidf}. Latent semantic analysis (LSA) uses singular-value decomposition to project sparse representations into a lower-dimensional dense space intended to capture higher-order associations \citep{deerwester1990lsa}. Modern dense retrieval replaces this decomposition with learned neural encoders and can outperform sparse retrieval when representation learning and supervision are sufficient \citep{karpukhin2020dpr}.

Biomedical language models provide domain-specific representations that are stronger than generic data-derived decompositions for many clinical and biomedical tasks. BioClinicalBERT was pretrained on clinical text \citep{alsentzer2019clinicalbert}, while SapBERT explicitly aligns biomedical entity representations using UMLS synonym structure \citep{liu2021sapbert}. These models are not used in the present experiment because the objective is to establish a reproducible baseline using only the challenge data and standard local machine-learning components. They are important next baselines for terminology-backed retrieval.

\subsection{Retrieval augmentation and ontology-grounded clinical mapping}
RAG combines parametric models with non-parametric retrieved evidence \citep{lewis2020rag}. In biomedicine, systematic evidence suggests that RAG can improve multiple language-model applications, while also showing that retrieval design, source quality, and clinical evaluation remain critical \citep{liu2025ragreview}. Ontology-grounded RAG extends this idea by using formal domain structure to guide retrieval rather than relying only on unstructured passages \citep{sharma2025ograg}.

SNOMED CT mapping is beginning to adopt this pattern directly. Huh evaluated GPT-4-based strategies, including retrieval augmentation, for mapping local medical terminology to SNOMED CT \citep{huh2025mapping}. Ehrler and colleagues described a hybrid pipeline in which embedding-based candidate retrieval is followed by LLM disambiguation for semantic mapping to SNOMED CT \citep{ehrler2026rag}. These studies motivate candidate-retrieval plus reranking architectures. The present study isolates an earlier question: whether adding retrieved note evidence improves concept ranking before introducing external terminology knowledge or a generative model.

\subsection{Relationship to our previous semantic analysis}
Our earlier MIMIC-IV study compared note-level SNOMED CT co-occurrence with embedding-based semantic similarity and found that the two signals were related but non-redundant \citep{noori2026semantic}. That work was descriptive and exploratory: it characterized concept relationships, temporal patterns, clusters, and embedding-based suggestions. The present study changes the endpoint from semantic characterization to held-out recommendation. Gold target mentions are removed, candidate concepts are ranked, and performance is measured with retrieval metrics and uncertainty estimates. This distinction is important because a relationship that appears clinically plausible in exploratory analysis may not improve out-of-sample ranking.

\section{Data}
\subsection{Source}
We use the SNOMED CT Entity Linking Challenge v1.2.1 resource on PhysioNet \citep{hardman2026challenge}, derived from MIMIC-IV-Note v2.2 \citep{johnson2023mimicnote}. MIMIC-IV-Note contains deidentified clinical free text under credentialed access. The challenge release consists of 272 annotated discharge summaries and approximately 75,000 annotations; the release documentation states that annotations were created by medically trained annotators, with substantial double annotation and adjudication \citep{hardman2026challenge}.

The v1.2.1 archive analyzed here contains exactly 75,491 annotation rows. Its annotation table retains labels identifying the historical \texttt{train} and \texttt{test} partitions and rows marked \texttt{proposed\_\allowbreak ACCEPTED}. We assign each accepted quality-control addition to the historical partition of its note. This produces 52,078 annotations from 204 training notes and 23,413 annotations from 68 test notes. No test note is used to fit text vectorizers, concept prototypes, latent semantic projections, or retrieval indexes.

\begin{table}[H]
\centering
\caption{Dataset statistics after assigning accepted quality-control additions to their note partition. Repeated mentions are retained in the annotation counts but collapsed for note-concept recommendation pairs.}
\label{tab:data}
\begin{tabular}{lrr}
\toprule
Metric & Training & Historical test \\
\midrule
Clinical notes & 204 & 68 \\
Annotation rows & 52,078 & 23,413 \\
Unique SNOMED CT concepts & 5,317 & 4,068 \\
Unique note-concept pairs & 32,591 & 14,179 \\
Mean unique concepts per note & 159.8 & 208.5 \\
Median unique concepts per note & 155.5 & 202.0 \\
IQR unique concepts per note & 123.8--192.3 & 169.0--243.3 \\
\bottomrule
\end{tabular}
\end{table}

The historical test notes are concept-dense relative to training notes: they contain 208.5 unique concepts per note on average versus 159.8 in training. This distribution difference increases the difficulty of recovering a single masked concept from a crowded clinical context.

\subsection{Vocabulary overlap and the long tail}
The training partition contains 5,317 unique concept IDs and the test partition contains 4,068. Only 2,790 concept IDs occur in both partitions. At the type level, 1,278 of the 4,068 test concepts (31.42\%) are absent from the training vocabulary. Because frequently occurring concepts contribute more note-concept pairs, instance-level coverage is higher: 12,809 of 14,179 test pairs (90.34\%) have a target concept seen during training, while 1,370 pairs (9.66\%) are unseen.

The training vocabulary is strongly long-tailed. More than half of training concepts (2,768; 52.06\%) appear in exactly one training note, and 3,533 concepts (66.45\%) appear in at most two. The median concept appears in one note and the upper quartile begins at four notes. This extreme sparsity is central to interpreting retrieval performance.

\begin{figure}[t]
\centering
\includegraphics[width=0.72\textwidth]{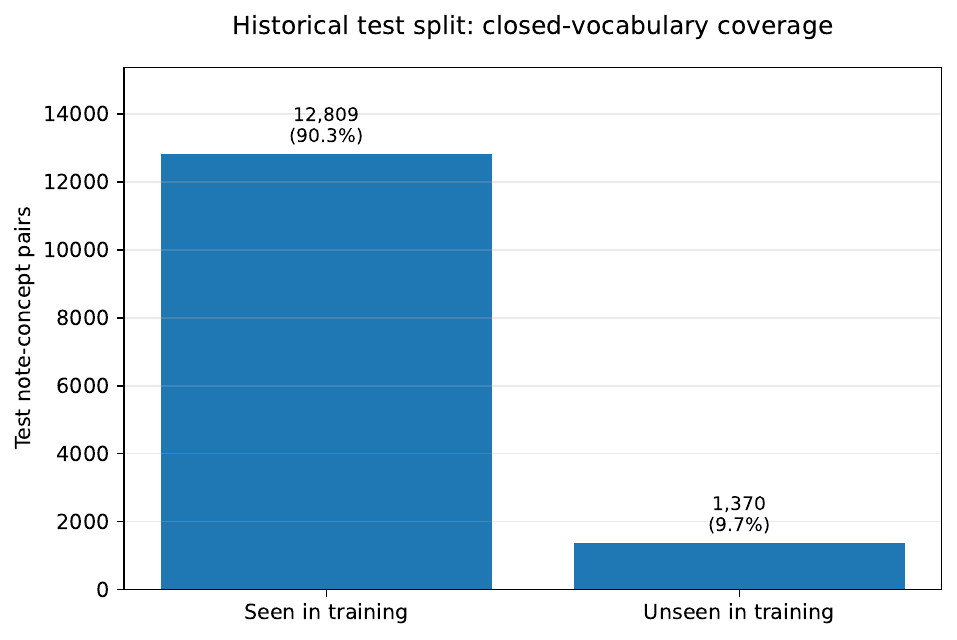}
\caption{Closed-vocabulary coverage of historical test note-concept pairs. Unseen targets cannot be produced by a recommender whose candidate inventory contains only concepts observed in training.}
\label{fig:coverage}
\end{figure}

\section{Masked-Concept Recommendation Task}
\subsection{Instance construction}
We create one recommendation instance for each unique pair of note ID and SNOMED CT concept ID. Repeated mentions of the same concept in the same note are therefore collapsed. The first annotated mention serves as the anchor location. A local context window extends 350 characters to the left of the anchor start and 350 characters to the right of the anchor end, clipped at note boundaries.

To remove direct lexical evidence, every annotated span of the target concept that overlaps this local window is replaced with the token \texttt{[MASK]}. The target surface form is never used as a model feature. A synthetic example is:

\begin{quote}
\small Original: ``The patient developed acute kidney injury after surgery and creatinine increased.''\\
Masked query: ``The patient developed \texttt{[MASK]} after surgery and creatinine increased.''
\end{quote}

This task is intentionally harder than entity linking because the most obvious concept string is removed. It asks whether context alone is sufficient to recover the missing concept.

\subsection{Candidate inventory and evaluation scope}
The candidate inventory is the 5,317 concepts observed in the historical training partition. Primary ranking metrics are computed on the 12,809 test pairs whose target concept is in this inventory. We call this the \emph{closed-vocabulary} evaluation. Unseen targets are reported separately rather than silently removed. For reference, if every unseen pair is counted as a failure, the strongest model's Recall@10 falls from 33.43\% to 30.20\%, and its MRR falls from 0.2114 to 0.1910.

This choice reflects what the challenge archive alone can support. A full SNOMED CT system should retrieve from a licensed terminology release containing concept descriptions, synonyms, and relations, enabling prediction of concepts never observed in the annotated notes. The challenge archive alone does not provide that candidate knowledge base.

\begin{figure}[t]
\centering
\includegraphics[width=0.96\textwidth]{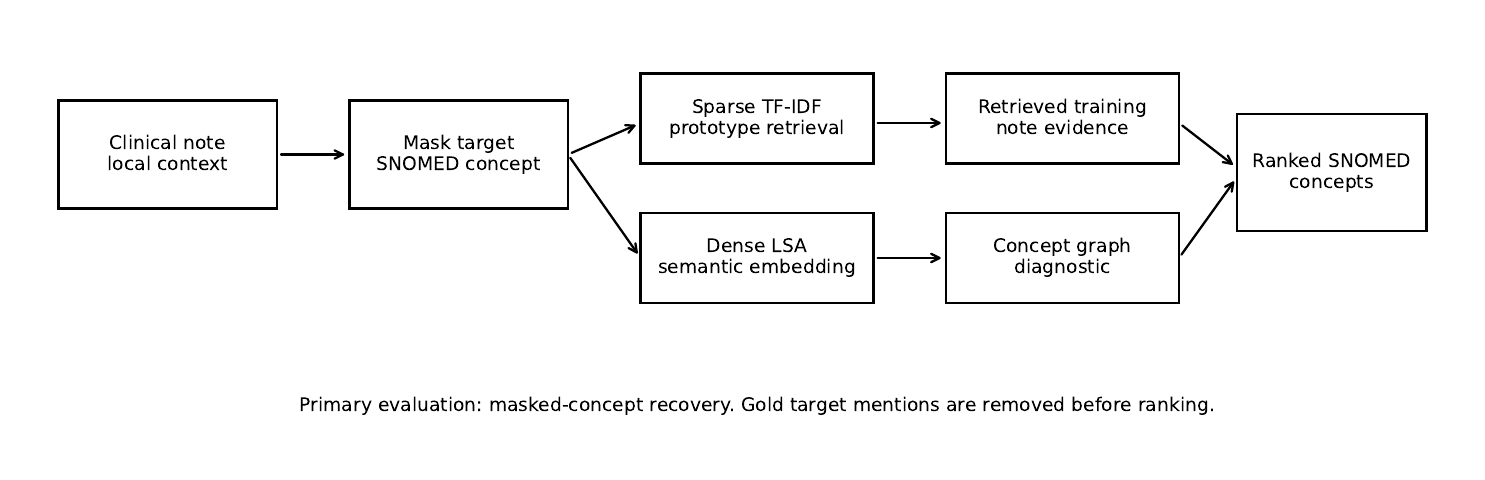}
\caption{Evaluation framework. The primary task removes the target mention and compares sparse, dense, and retrieval-derived ranking signals. The concept-graph condition uses gold remaining concepts only as a diagnostic and is not treated as an end-to-end system.}
\label{fig:framework}
\end{figure}

\section{Methods}
\subsection{Popularity baseline}
The popularity baseline ranks candidates by the number of training notes in which each concept occurs. This quantifies how much performance can be obtained from the marginal concept distribution without using query text.

\subsection{Sparse TF-IDF concept prototypes}
Masked training contexts are represented with a word-level TF-IDF vectorizer using unigrams and bigrams, sublinear term frequency, a minimum document frequency of two, and a maximum vocabulary of 50,000 features. TF-IDF is a standard and interpretable retrieval representation \citep{salton1988tfidf}.

For concept $c$, let $Q_c$ be the set of masked training contexts whose target is $c$. The sparse prototype is the normalized mean
\begin{equation}
\mathbf{p}_c = \frac{1}{|Q_c|}\sum_{q \in Q_c}\mathbf{x}_q,
\end{equation}
where $\mathbf{x}_q$ is the L2-normalized TF-IDF vector. A test query is ranked by cosine similarity to all concept prototypes. Because target strings are masked in both training and test contexts, this is context-based retrieval rather than surface-form matching.

\subsection{Dense LSA embeddings}
We apply truncated singular-value decomposition to the training TF-IDF matrix and retain 96 latent components, producing a data-derived dense semantic representation \citep{deerwester1990lsa}. The fitted projection explains 18.57\% of TF-IDF variance. Training context embeddings are L2-normalized and averaged per concept to create dense concept centroids. Test contexts are projected with the same fitted transformation and ranked by cosine similarity to concept centroids.

This LSA representation should not be interpreted as equivalent to a pretrained biomedical encoder. Rather, it provides a controlled dense baseline learned only from the challenge data. ClinicalBERT- or SapBERT-based retrieval is reserved for future work \citep{alsentzer2019clinicalbert,liu2021sapbert}.

\subsection{Sparse-dense hybrid}
An internal note-level validation split is created from the historical training notes. A sparse weight of 0.75 and dense weight of 0.25 is selected from a small mixture grid using validation MRR. After weight selection, representations are refit on all 204 historical training notes. For a query $q$ and concept $c$,
\begin{equation}
s_{SD}(q,c)=0.75\,\tilde{s}_{TFIDF}(q,c)+0.25\,\tilde{s}_{LSA}(q,c),
\end{equation}
where each score family is min-max normalized within the query before fusion.

\subsection{Training-note retrieval evidence}
To approximate the retrieval component of a RAG pipeline using only the challenge dataset, each masked query is compared with all 204 historical training notes in the fitted TF-IDF space. The ten most similar training notes are retained. A candidate receives the similarity-weighted sum of evidence from retrieved notes in which that concept is annotated. Scores are min-max normalized per query.

This model provides non-parametric memory in the RAG sense: predictions can be supported by retrieved training documents rather than only a concept prototype. However, there is no generative model in this experiment, so we refer to it as \emph{retrieved-note evidence} rather than generative RAG.

\subsection{Retrieval-augmented hybrid}
The retrieval-augmented condition combines 85\% of the sparse-dense score with 15\% retrieved-note evidence:
\begin{equation}
s_{RA}(q,c)=0.85\,s_{SD}(q,c)+0.15\,\tilde{s}_{note}(q,c).
\end{equation}
Internal validation favored zero retrieval weight overall; the 15\% condition is retained as a sensitivity analysis to directly test whether a modest retrieval contribution improves historical test performance.

\subsection{Concept-graph diagnostic}
We additionally build a training-only concept co-occurrence graph from note-level binary concept incidence. Concept-concept edge weights are cosine similarity between their training-note occurrence vectors. For each test target, all other gold concepts in that test note are treated as known seed concepts, while the target itself is removed. Candidate graph scores are the summed similarities to the seed set.

This is explicitly an \emph{oracle diagnostic}: it assumes perfect knowledge of all other annotated concepts in the note. It is included because our prior work found that co-occurrence and embedding similarity contain complementary information \citep{noori2026semantic}. It should not be compared to text-only models as an end-to-end deployment condition. We also report a fixed mixture of 65\% sparse-dense text score and 35\% oracle graph score to test whether even privileged graph context can materially change the ranking.

\subsection{Implementation details and reproducibility}
All preprocessing, feature fitting, prototype construction, latent projections, retrieval indexes, and graph statistics are learned from the historical training partition only. The random seed is fixed at 42. Table~\ref{tab:hyper} summarizes the principal analysis settings. The sparse-dense mixture is selected on an internal note-level validation split. Validation favored a retrieval weight of zero; therefore, the 15\% retrieved-note contribution is reported as a prespecified sensitivity condition rather than as a test-set-optimized setting.

\begin{table}[H]
\centering
\caption{Principal analysis settings.}
\label{tab:hyper}
\small
\begin{tabularx}{0.93\textwidth}{lY}
\toprule
Setting & Value \\
\midrule
Masked context window & 350 characters on each side of the anchor mention \\
TF-IDF features & Word unigrams and bigrams; sublinear TF; minimum document frequency 2; maximum document frequency 0.995; maximum 50,000 features \\
Dense representation & Truncated SVD / LSA with 96 components; random seed 42 \\
Sparse-dense fusion & 0.75 sparse + 0.25 dense, selected using internal note-level validation MRR \\
Retrieved evidence & Top 10 historical training notes ranked by TF-IDF cosine similarity \\
Retrieval sensitivity weight & 0.15 retrieved-note evidence + 0.85 sparse-dense score \\
Oracle graph mixture & 0.35 graph + 0.65 sparse-dense score; diagnostic only \\
Uncertainty estimation & 2,000 paired bootstrap resamples clustered by test note \\
\bottomrule
\end{tabularx}
\end{table}

\section{Evaluation}
For each closed-vocabulary test instance, all 5,317 concepts observed in the historical training partition are ranked. We report Recall@1, Recall@5, Recall@10, mean reciprocal rank (MRR), and normalized discounted cumulative gain at 10 (nDCG@10). With one gold target per query, these metrics measure complementary aspects of whether the correct concept appears near the top of the ranked list. Consistent with our earlier work emphasizing robust evaluation of automated ontology annotation systems \citep{noori2025metrics}, we also report vocabulary coverage, frequency-stratified performance, median rank, and clustered uncertainty rather than relying on a single aggregate metric.

Targets are stratified by the number of historical training notes in which the concept occurs: rare (1--2 notes), medium (3--10 notes), and frequent (more than 10 notes). The closed-vocabulary test set contains 1,641 rare, 2,739 medium-frequency, and 8,429 frequent target instances. These groups sum to the 12,809 evaluable test pairs.

For statistical comparison, we use 2,000 paired bootstrap resamples clustered by test note. Resampling at the note level preserves dependence among the many masked-concept instances derived from the same discharge summary. We report percentile 95\% confidence intervals for the strongest sparse model and for paired metric differences. All test instances remain untouched during feature fitting and parameter selection.

\section{Results}
\subsection{Overall recommendation performance}
Table~\ref{tab:results} presents closed-vocabulary results on 12,809 historical test note-concept pairs. Sparse TF-IDF concept prototypes provide the strongest overall performance, reaching 14.81\% Recall@1, 33.43\% Recall@10, 0.2114 MRR, and 0.2297 nDCG@10. The 95\% note-cluster bootstrap interval is 31.87\%--34.87\% for Recall@10 and 0.2000--0.2227 for MRR.

Dense LSA embeddings are substantially weaker, with 17.10\% Recall@10 and 0.0735 MRR. The sparse-dense mixture also underperforms sparse retrieval alone, indicating that the 96-dimensional latent representation removes discriminative lexical structure needed for this low-resource masked-context task.

\begin{table}[H]
\centering
\caption{Closed-vocabulary masked-concept recommendation performance on 12,809 historical test note-concept pairs. The oracle graph rows use gold remaining concepts and are diagnostic rather than deployable end-to-end systems. Best non-oracle value in each primary metric is bold.}
\label{tab:results}
\small
\begin{tabular}{lrrrrr}
\toprule
Model & R@1 & R@5 & R@10 & MRR & nDCG@10 \\
\midrule
Popularity & 0.0052 & 0.0254 & 0.0498 & 0.0250 & 0.0228 \\
Sparse TF-IDF prototype & \textbf{0.1481} & \textbf{0.2603} & \textbf{0.3343} & \textbf{0.2114} & \textbf{0.2297} \\
Dense LSA embedding & 0.0255 & 0.0989 & 0.1710 & 0.0735 & 0.0849 \\
Sparse + dense & 0.1364 & 0.2465 & 0.3197 & 0.1983 & 0.2162 \\
Retrieved-note evidence & 0.0087 & 0.0456 & 0.0849 & 0.0379 & 0.0389 \\
Retrieval-augmented hybrid & 0.1301 & 0.2452 & 0.3199 & 0.1937 & 0.2125 \\
\midrule
Oracle concept-graph & 0.0166 & 0.0618 & 0.0685 & 0.0442 & 0.0430 \\
Hybrid + oracle graph & 0.1465 & 0.2498 & 0.3227 & 0.2048 & 0.2221 \\
\bottomrule
\end{tabular}
\end{table}

\begin{figure}[t]
\centering
\includegraphics[width=0.86\textwidth]{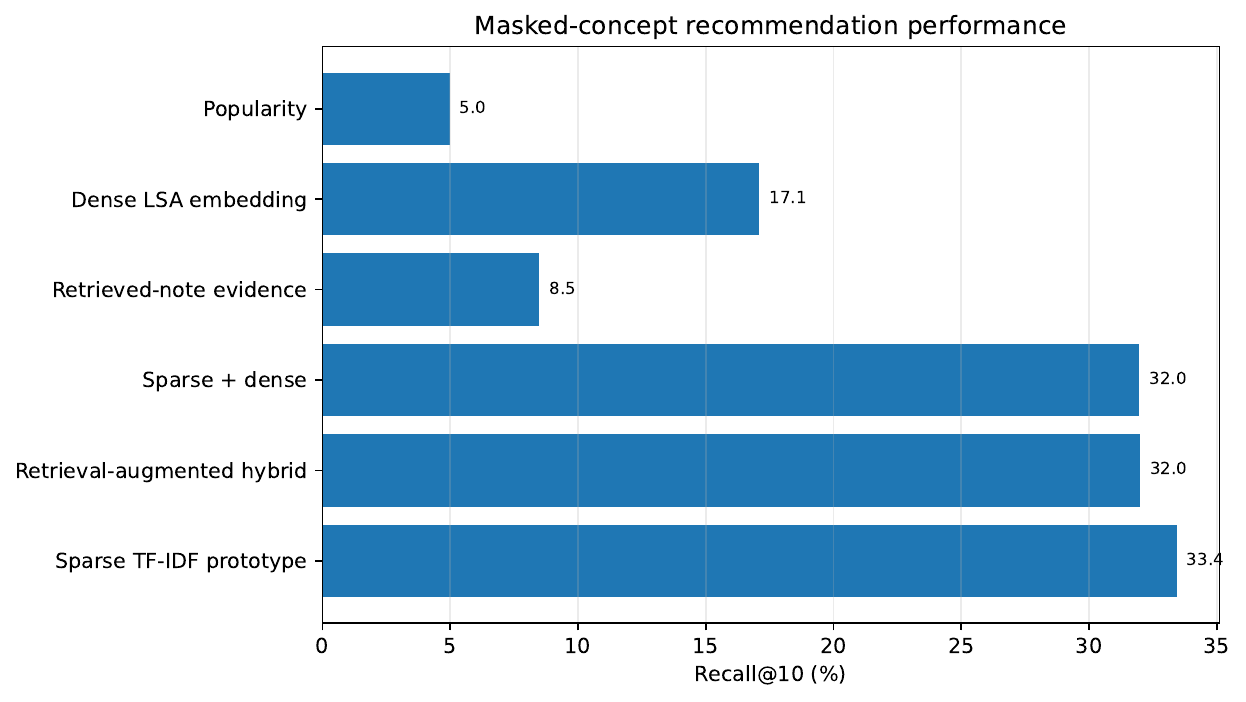}
\caption{Recall@10 for the principal text-only systems. Sparse TF-IDF prototypes outperform the data-derived dense representation and retrieved-note evidence.}
\label{fig:r10}
\end{figure}

\subsection{Does retrieval augmentation help?}
Retrieved-note evidence alone performs poorly (8.49\% Recall@10), showing that document-level similarity across only 204 training notes is too coarse for fine-grained concept recovery. Adding 15\% retrieval evidence to the sparse-dense hybrid changes Recall@10 only from 31.97\% to 31.99\% but lowers MRR from 0.1983 to 0.1937.

More importantly, the strongest sparse baseline remains significantly better than the retrieval-augmented hybrid. The paired note-cluster bootstrap difference (retrieval-augmented minus sparse) is -0.0144 in Recall@10, with a 95\% confidence interval from -0.0184 to -0.0104. The MRR difference is -0.0177, with a 95\% interval from -0.0212 to -0.0142. Thus, under this retrieval design, augmentation is associated with lower ranking performance than the sparse baseline.

The oracle concept-graph diagnostic also does not solve the task by itself. Its Recall@10 is only 6.85\%. Combining graph and text signals reaches 32.27\% Recall@10, still below sparse TF-IDF. The graph mixture changes MRR by -0.0066 relative to sparse TF-IDF, and its paired 95\% interval (-0.0140 to 0.0009) includes zero. These results suggest that note-level co-occurrence structure is too broad for recovering a specific concept from a small masked context, even when the other note concepts are known perfectly.

\begin{table}[H]
\centering
\caption{Paired note-cluster bootstrap comparisons against the sparse TF-IDF prototype. Differences are comparator minus sparse TF-IDF; negative values favor the sparse model.}
\label{tab:bootstrap}
\small
\begin{tabular}{lrrr}
\toprule
Comparator & Metric & Observed difference & 95\% CI \\
\midrule
Sparse + dense & Recall@10 & -0.0146 & [-0.0182, -0.0111] \\
Sparse + dense & MRR & -0.0131 & [-0.0147, -0.0116] \\
Retrieval-augmented hybrid & Recall@10 & -0.0144 & [-0.0184, -0.0104] \\
Retrieval-augmented hybrid & MRR & -0.0177 & [-0.0212, -0.0142] \\
Hybrid + oracle graph & Recall@10 & -0.0116 & [-0.0196, -0.0035] \\
Hybrid + oracle graph & MRR & -0.0066 & [-0.0140, 0.0009] \\
\bottomrule
\end{tabular}
\end{table}

\subsection{Long-tail performance}
Concept frequency strongly determines performance (Table~\ref{tab:freq}). For sparse TF-IDF, Recall@10 is 7.74\% for targets appearing in only one or two training notes, 16.61\% for targets appearing in three to ten notes, and 43.90\% for targets appearing in more than ten notes. The corresponding median ranks are 991, 174, and 14. Dense LSA shows the same qualitative pattern and remains below the sparse model in every frequency stratum.

\begin{table}[H]
\centering
\caption{Recall@10 by target frequency in historical training notes. The $N$ column gives the number of closed-vocabulary test instances in each stratum.}
\label{tab:freq}
\small
\begin{tabular}{lrrrr}
\toprule
Frequency stratum & $N$ & Sparse TF-IDF & Dense LSA & Retrieval-augmented \\
\midrule
Rare (1--2 notes) & 1,641 & \textbf{0.0774} & 0.0622 & 0.0695 \\
Medium (3--10 notes) & 2,739 & \textbf{0.1661} & 0.1395 & 0.1548 \\
Frequent ($>$10 notes) & 8,429 & \textbf{0.4390} & 0.2024 & 0.4224 \\
\bottomrule
\end{tabular}
\end{table}

\begin{figure}[t]
\centering
\includegraphics[width=0.87\textwidth]{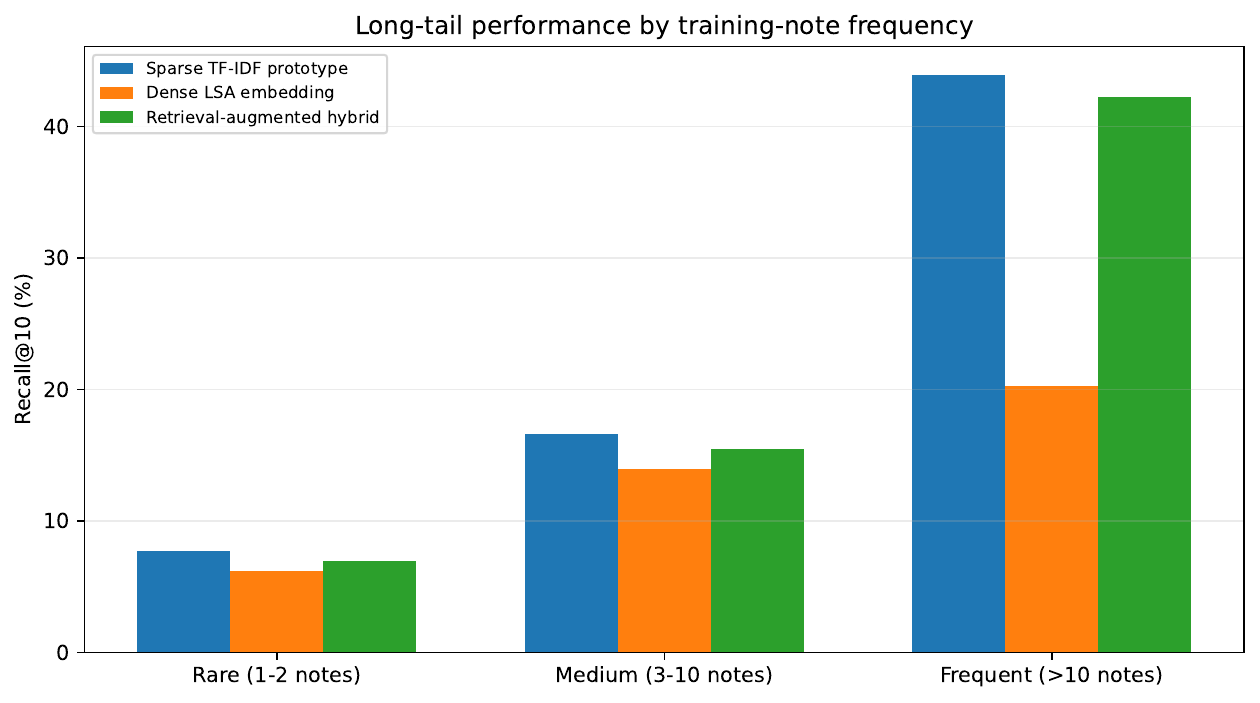}
\caption{Long-tail effect. Recall@10 increases sharply with the number of training notes containing the target concept.}
\label{fig:freq}
\end{figure}

\section{Discussion}
\subsection{RQ1: Sparse local context is the strongest baseline}
The first research question asked how well sparse and dense representations recover a masked concept. Sparse TF-IDF concept prototypes are clearly strongest among the evaluated text-only approaches, reaching 33.43\% Recall@10 and 0.2114 MRR. Dense LSA reaches 17.10\% Recall@10, and sparse-dense fusion does not recover the lost performance. The result is plausible given the data regime: only 204 training notes support 5,317 candidate concepts, and more than half of those concepts appear in exactly one training note.

In this setting, a 96-dimensional latent projection compresses many distinct clinical contexts, whereas TF-IDF preserves specific neighboring words, medication names, measurements, section phrases, and procedural language. The LSA projection explains 18.57\% of TF-IDF variance, which is consistent with substantial compression. These findings should not be interpreted as evidence against neural biomedical embeddings. LSA is a controlled data-derived dense baseline, not a substitute for ClinicalBERT, SapBERT, or a terminology-trained bi-encoder \citep{alsentzer2019clinicalbert,liu2021sapbert}.

\subsection{RQ2: Whole-note retrieval does not improve ranking}
The second research question asked whether retrieved evidence improves concept recommendation. Retrieved-note evidence alone reaches only 8.49\% Recall@10. Adding a 15\% retrieval contribution to the sparse-dense score yields 31.99\% Recall@10 and 0.1937 MRR, both below the sparse TF-IDF prototype. The paired bootstrap intervals in Table~\ref{tab:bootstrap} exclude zero for both Recall@10 and MRR, indicating that the reduction is systematic across resampled test notes rather than an isolated aggregate fluctuation.

The likely problem is granularity. A masked local context may concern one finding, procedure, or measurement, while a discharge summary contains hundreds of concepts. Retrieving entire training notes therefore introduces many concepts unrelated to the missing target. This suggests that retrieval should operate over smaller evidence units such as sentences, note sections, mention-centered passages, or terminology records. It also supports a distinction between \emph{retrieval augmentation} and \emph{generative RAG}: a generative model cannot compensate for weak or unfocused candidate evidence if the relevant concept is not retrieved.

\subsection{RQ3: The terminology long tail dominates performance}
The third research question concerns frequency and vocabulary coverage. The largest performance difference is not between sparse and dense representations but between frequent and rare target concepts. Sparse Recall@10 increases from 7.74\% for concepts observed in one or two training notes to 43.90\% for concepts observed in more than ten. The median rank improves from 991 to 14 across the same frequency groups. Because 66.45\% of the training vocabulary appears in at most two notes, most candidate concepts receive very little supervised contextual evidence.

Coverage creates an additional limitation. Of 4,068 unique concept types in the historical test partition, 1,278 (31.42\%) do not occur in training. At the instance level, 1,370 of 14,179 test note-concept pairs (9.66\%) are therefore impossible for a closed-vocabulary recommender. When these unseen targets are counted as failures, sparse TF-IDF Recall@10 falls from 33.43\% to 30.20\% and MRR from 0.2114 to 0.1910. A terminology-backed candidate index is therefore necessary for genuine zero-shot recommendation.

\subsection{Implications for ontology-grounded RAG}
The negative retrieval result is informative for the next system design. Rather than retrieving whole discharge summaries and adding their concept counts to a ranking score, a stronger architecture should index the terminology itself. Each SNOMED CT concept can be represented using its preferred term, fully specified name, synonyms, hierarchy, and selected relations. SapBERT or another biomedical entity encoder can provide concept embeddings \citep{liu2021sapbert}, while a clinical encoder can represent the masked note context \citep{alsentzer2019clinicalbert}. Fine-grained note passages and terminology records can then be retrieved jointly, after which a constrained language model can rerank only valid candidate concept IDs.

This direction extends our earlier research trajectory. The Bi-GRU study addressed supervised concept-span annotation \citep{noori2025bigru}; the ontology-annotation metrics study emphasized careful evaluation of automated annotation systems \citep{noori2025metrics}; and our semantic analysis showed that co-occurrence and embedding similarity encode complementary clinical relationships \citep{noori2026semantic}. The present results add an out-of-sample ranking perspective: co-occurrence and retrieval signals may be clinically meaningful without necessarily improving masked-concept recommendation when used at note level. Ontology-grounded retrieval may be most valuable for rare and unseen concepts because it introduces knowledge unavailable from the 204 training notes alone \citep{sharma2025ograg}.

\subsection{Scope of the retrieval-augmented evaluation}
The present experiment evaluates retrieval augmentation at the ranking layer, not a complete generative RAG system. This scope is deliberate. The analyzed challenge archive contains clinical notes, span annotations, and concept identifiers, but it does not provide the full SNOMED CT terminology release required to construct a licensed ontology-backed knowledge base. It would therefore be inappropriate to attribute the observed results to LLM generation or to claim zero-shot ontology-grounded retrieval. Instead, the study establishes a transparent baseline against which those additions can later be measured.

\section{Limitations}
Several limitations should guide interpretation. First, masked-concept recovery is a constructed retrospective task rather than the original challenge endpoint. It measures context-based recommendation after removing a known gold mention and should not be compared directly with official span-detection or character-level challenge scores \citep{davidson2025challenge}.

Second, one instance is created per note-concept pair using the first annotated mention as the anchor. Other mentions of the same concept may occur in more or less informative contexts. Section-aware sampling, multiple anchors, and multiple context-window sizes should be evaluated in future work.

Third, the dense representation is LSA trained only on the available challenge contexts. This design isolates what can be learned from the supplied data but limits conclusions about neural semantic retrieval. Biomedical encoders such as ClinicalBERT and SapBERT should be evaluated directly \citep{alsentzer2019clinicalbert,liu2021sapbert}.

Fourth, the retrieval corpus contains only 204 training discharge summaries and uses whole notes as evidence units. Fine-grained passage retrieval may be substantially stronger. Fifth, the candidate inventory contains only concepts seen during training, leaving 9.66\% of test instances out of vocabulary. A complete terminology index is required to evaluate unseen-concept recommendation. Sixth, the concept-graph condition assumes gold knowledge of all remaining concepts in the test note and is therefore an oracle diagnostic rather than a deployable end-to-end model. Finally, the study uses MIMIC-IV discharge documentation from one source and may not generalize to other institutions, note types, specialties, or real-time documentation workflows.

\section{Conclusion}
We evaluated SNOMED CT concept recommendation from masked clinical context using the SNOMED CT Entity Linking Challenge v1.2.1 release. Among 12,809 closed-vocabulary historical test pairs, sparse TF-IDF context prototypes achieved the best performance, with 33.43\% Recall@10 and 0.2114 MRR. Dense LSA, whole-note retrieval evidence, sparse-dense fusion, and simple retrieval augmentation did not improve this baseline. The retrieval-augmented hybrid reached 31.99\% Recall@10 and 0.1937 MRR, with paired note-cluster bootstrap intervals showing lower performance than sparse TF-IDF.

The dominant challenge is terminology sparsity. Recall@10 is only 7.74\% for concepts represented in one or two training notes, and 9.66\% of test note-concept pairs contain concepts absent from the training vocabulary. These findings suggest that the next improvement should come from expanding the candidate knowledge base rather than increasing whole-note retrieval weight. A full follow-up system should index SNOMED CT descriptions, synonyms, and relations; use biomedical entity and clinical-context encoders for candidate retrieval; retrieve evidence at finer granularity; and constrain any language-model reranker to valid candidate identifiers. The benchmark and reproducible baselines reported here provide a measured reference point for testing whether those ontology-grounded additions improve recommendation quality.

\section*{Data and Code Availability}
The clinical notes and annotations are credentialed and are not redistributed with this manuscript. They are available through the SNOMED CT Entity Linking Challenge v1.2.1 resource on PhysioNet, subject to its access requirements and data use agreement \citep{hardman2026challenge}. The accompanying reproducibility package contains analysis and figure-generation code, aggregate result tables, dataset statistics, and bootstrap summaries. The scripts expect local copies of \texttt{train\_notes.csv} and \texttt{train\_annotations.csv}. No raw clinical text is included in the arXiv source package or reproducibility archive.

\section*{Ethics Statement}
This study is a secondary analysis of deidentified MIMIC-IV-Note data accessed under PhysioNet credentialing. No new patient contact, intervention, or prospective clinical decision support was performed. The outputs are research recommendations and should not be interpreted as clinical advice.

\bibliographystyle{unsrtnat}
\bibliography{references}
\end{document}